\documentclass[journal,12pt,onecolumn,draftcls]{IEEEtran}
\usepackage[left=1.24in, right=1.24in, bottom=1in, top=1in]{geometry}

\usepackage{cite}
\usepackage{todonotes}
\usepackage{amsmath,amssymb,amsfonts}
\usepackage{algorithmic}
\usepackage{graphicx}
\usepackage{textcomp}
\usepackage{xcolor}
\usepackage{tabularx}
\usepackage{booktabs}
\usepackage{multirow}
\usepackage{collcell}
\usepackage{hhline}
\usepackage{tikz}
\usepackage[normalem]{ulem}
\usepackage{pgfplots,etoolbox}
\usepackage{pgfplotstable}
\usepackage{pgffor}
\usepackage{ifthen}
\usepackage{float}
\usepackage{vcell}
\usepackage[hidelinks]{hyperref}

\pgfplotsset{compat=1.18}
\graphicspath{ {./figures/} }
\date{May 2023}

\begin{document}

\title{Synthetic Image Detection:\\
       Highlights from the IEEE Video and Image Processing Cup 2022 Student Competition}

\author{D. Cozzolino, K. Nagano, L. Thomaz, A. Majumdar, and L. Verdoliva}

\maketitle

The Video and Image Processing (VIP) Cup is a student competition that takes place each year at the IEEE International Conference on Image Processing.
The 2022 IEEE VIP Cup asked undergraduate students to develop a system capable of distinguishing pristine images from generated ones.
The interest in this topic stems from the incredible advances in the AI-based generation of visual data, with tools that allows the synthesis of highly realistic images and videos.
While this opens up a large number of new opportunities, it also undermines the trustworthiness of media content and fosters the spread of disinformation on the internet. 
Recently there was strong concern about the generation of extremely realistic images by means of editing software that includes the recent technology on diffusion models \cite{theguardian, theverge}.
In this context, there is a need to develop robust and automatic tools for synthetic image detection.

In the literature, there has been an intense research effort to develop effective forensic image detectors, and many of them, if properly trained, appear to provide excellent results \cite{Verdoliva2020media}.
Such results, however, usually refer to ideal conditions and rarely stand the challenge of real-world application.
First of all, testing a detector on images generated by the very same models seen in the training phase, leads to overly optimistic results.
In fact, this is not a realistic scenario.
With the evolution of technology, new architectures and different ways of generating synthetic data are continuously proposed \cite{karras2019style, nichol2021glide, Dayma_DALLE_Mini_2021,rombach2022high,balaji2022ediff}.
Therefore, detectors trained on some specific sources will end up working on target data of a very different nature, often with disappointing results.
In these conditions, the ability of generalizing to new data becomes crucial to keep providing a reliable service.
Moreover, detectors are often required to work on data that have been seriously impaired in several ways.
For example, when images are uploaded on social networks, they are normally resized and compressed to meet internal constraints. 
These operations tend to destroy important forensic traces, calling for detectors that are robust to such events and degrade performance gracefully.
To summarize, in order to operate successfully in the wild, a detector should be robust to image impairments and, at the same time, able to generalize well on images coming from diverse and new models.

In the scientific community there is still insufficient awareness (although growing) of the centrality of these aspects in the development of reliable detectors.
Therefore, we took the opportunity of this VIP Cup to push further along this direction.
In designing the challenge, we decided to consider an up-to-date, realistic setting with test data including
{\it  i)} both fully synthetic and partially manipulated images and
{\it ii)} images generated by both established GAN models and newer architectures, such as diffusion-based models.
With the first dichotomy, we ask that the detectors be robust to the occurrence of images that are only partially synthetic, thus with limited data on which to base the decision.
As for architectures, there is already a significant body of knowledge on the detection of GAN-generated images \cite{Gragnaniello2021GAN},
but new text-based diffusion models are now gaining the spotlight and generalization becomes the central issue.
With the 2022 IEEE VIP Cup we challenged teams to design solutions that are able to work in the wild 
as only a fraction of the generators used in the test data are known in advance. 

In this article, we present an overview of this challenge, including the competition setup, the teams and their technical approaches. Note that all the teams are composed by a professor, at most one graduate student (tutor), and undergraduate students (from a minimum of 3 to a maximum of 10 students).

\begin{figure}
    \centering
    \includegraphics[width=0.8\linewidth, clip, trim=20 180 20 0]{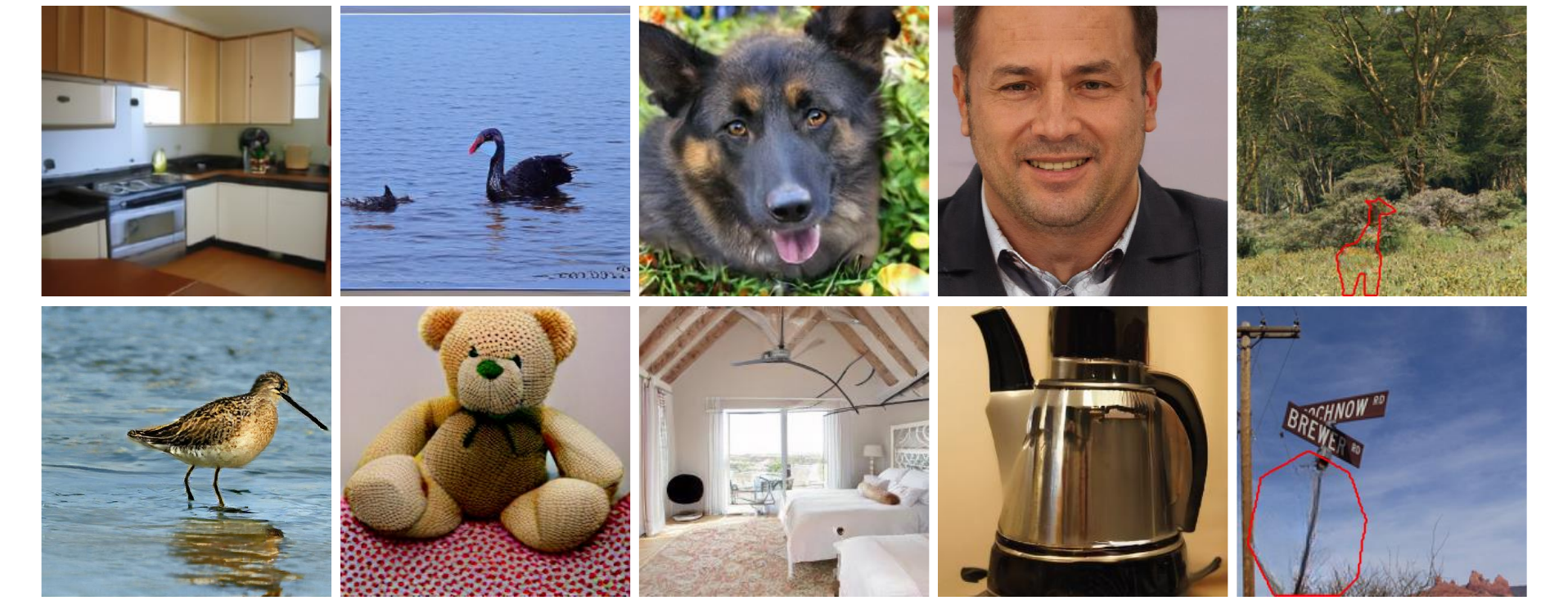}
    \caption{Examples of synthetic images from the datasets used in the open competition. 
    First row: samples from GLIDE \cite{nichol2021glide}, Taming Transformers \cite{esser2021taming}, StyleGAN2 \cite{karras2020analyzing}, StyleGAN3 \cite{karras2021alias}, inpainting with Gated Convolution \cite{yu2019free}. 
    Second row: samples from BigGAN \cite{brock2018large}, DALL-e mini \cite{Dayma_DALLE_Mini_2021}, ADM (Ablated Diffusion Model) \cite{dhariwal2021diffusion}, Latent Diffusion \cite{rombach2022high}, LaMa \cite{suvorov2022resolution}.
    Images in the last column are only locally manipulated (the regions outlined in red are synthetic).}
    \label{fig:examples}
\end{figure}

\section{Tasks, resources, and evaluation criteria}

{\bf Tasks.}
The challenge consists of two phases: an open competition (split into two parts) in which any eligible team can participate and an invitation-only final.
Phase 1 of the open competition is designed to provide teams with a simplified version of the problem at hand to familiarize themselves with the task, while Phase 2 is designed to tackle a more challenging task: synthetic data generated using architectures not present in the training. 
The synthetic images included in Phase 1 are generated using five known techniques, while the generated models used in Phase 2 are unknown.
During the final competition, the three highest scoring teams from the open competition are selected and are allowed to provide another submission graded on a new test set. 
Information about the challenge are also available at the link: \url{https://grip-unina.github.io/vipcup2022/}.

{\bf Resources.}
Participants were provided with a labelled training dataset of real and synthetic images. 
In particular, the dataset available for Phase 1 comprises real images from four datasets (FFHQ \cite{karras2019style}, Imagenet \cite{deng2009imagenet}, COCO \cite{lin2014microsoft}, LSUN \cite{yu2015lsun}), 
while synthetic images are generated using five known techniques: StyleGAN2 \cite{karras2020analyzing}, StyleGAN3 \cite{karras2021alias}, GLIDE \cite{nichol2021glide}, Taming Transformers \cite{esser2021taming}, and inpainted images with Gated Convolution \cite{yu2019free}.
All the images of the test data are randomly cropped and resized to 200$\times$200 pixels and then compressed using JPEG at different quality levels. 
This pipeline is used to simulate a realistic scenario where images are randomly resized and compressed as happens when they are uploaded to a social network. 
In addition, they all have the same dimensions to avoid leaking information on the used generators (some models only generate data at certain specific resolutions).
Some examples of generated images used during the competition are shown in Fig.\ref{fig:examples}.

Teams are provided with Python scripts to apply these same operations to the training dataset. 
For Phase 2, there were no available datasets since the generated models in this case are unknown to the teams.
However participants were free to use any external data, besides the competition data. 
In addition, participants were allowed to use any available state-of-the-art methods and algorithms to solve the problems of the challenge.

Teams were requested to provide the executable code to the organizers in order to test the algorithms on the evaluation datasets. 
The Python code has been executed inside a Docker container with a GPU of 16 GB with a time limit of one hour to process a total of 5000 images. 
The teams were allowed to submit their code and evaluate their performance five times during the period from August, 8th to September, 5th 2022.

{\bf Evaluation criteria.}
The submitted algorithms were scored by means of balanced accuracy for the detection task (Score = 0.7 $\times$ Accuracy-Phase-1 + 0.3 $\times$ Accuracy-Phase-2).
The three highest-scoring teams from the open competition stage were selected as finalists. 
These teams had the opportunity to make an additional submission on October 8th on a new dataset and were invited to compete in the final stage of the challenge at 2022 ICIP on October 16th, 2022 in Bordeaux. 
Due to some travel limitations, on that occasion, they could make a live or pre-recorded presentation, followed by a round of questions from a technical committee.
The event was hybrid to ensure a wide participation and allow teams who had VISA issues to attend virtually.
In the final phase of the challenge the judging committee considered the following parameters for the final evaluation (maximum score: 12 points):
\begin{itemize}
    \item the innovation of the technical solution (1-3 points),
    \item the performance achieved in phase 1 of the competition where only known models were used to generate synthetic data (1-3 points),
    \item the performance achieved in phase 2 of the competition where unknown models were used to generate synthetic data (1-3 points),
    \item the quality and clarity of the final report, a four pages full conference paper in the IEEE format (1-3 points),
    \item the quality and clarity of the final presentation (either pre-recorded or live) 15 minutes talk (1-3 points).
\end{itemize}

\section{2022 VIP Cup statistics and results}

The VIP Cup was run as an online class through the Piazza platform, which allowed to interact easily with the teams. 
In total, we received 82 registrations for the challenge, 26 teams accessed the Secure CMS platform and 13 teams made at least one valid submission. 
Teams were from 10 different countries across the world: Bangladesh, China, Germany, Greece, India, Italy, Poland, Sri Lanka, United States of America, and Vietnam.

\begin{figure}
    \centering
    \includegraphics[width=0.49\linewidth,page=1]{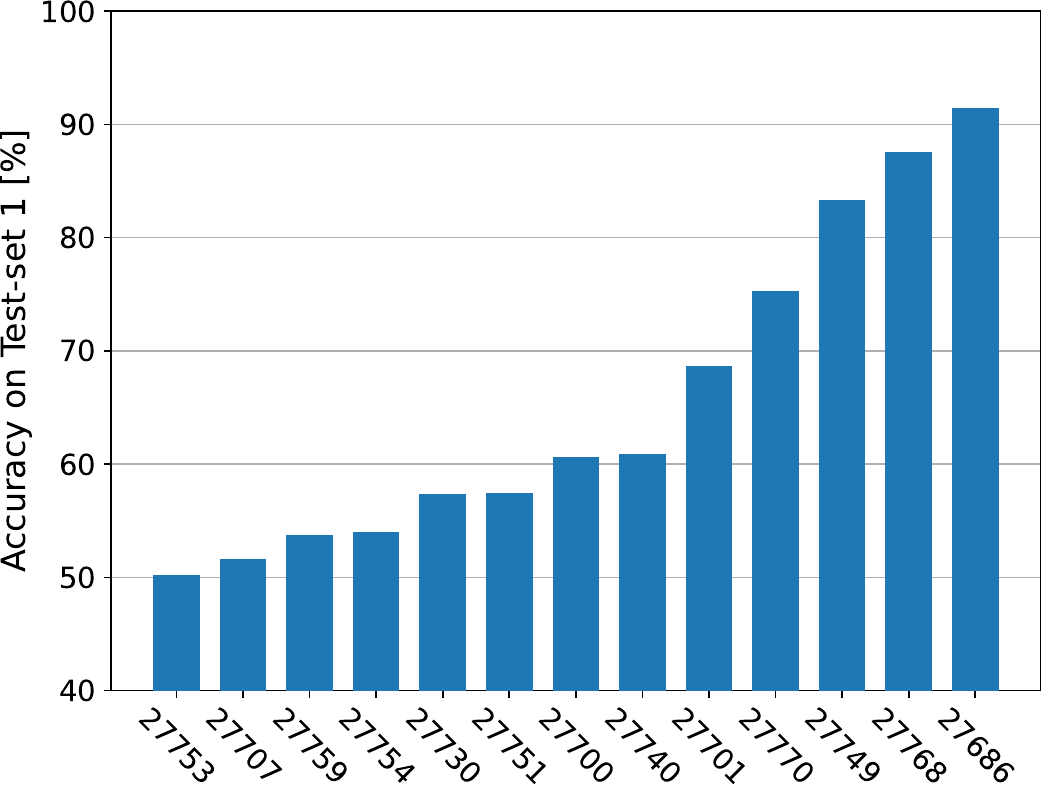}
    \includegraphics[width=0.49\linewidth,page=2]{fig2.pdf}
    \caption{The anonymized results in terms of accuracy of the 13 teams on the two open-competition datasets.}
    \label{fig:results_accuracy}
\end{figure}

Figure \ref{fig:results_accuracy} presents the accuracy results obtained by the 13 teams participating in the two phases of the open competition.
First, we can observe that the performance on the Test-set 1 including images from known generators are much higher than those obtained in an open set scenario, where generators are unknown. More specifically, accuracy drops of around 10\% for the best techniques, confirming the difficulty to detect synthetic images coming from unknown models. Then, we note that even for the simpler scenario only four teams were able to achieve and accuracy above 70\%, which highlights that designing a detector that can operate well both on fully and locally manipulated images is not an easy task.

\begin{figure}
    \centering
    \includegraphics[width=0.45\linewidth, page=1]{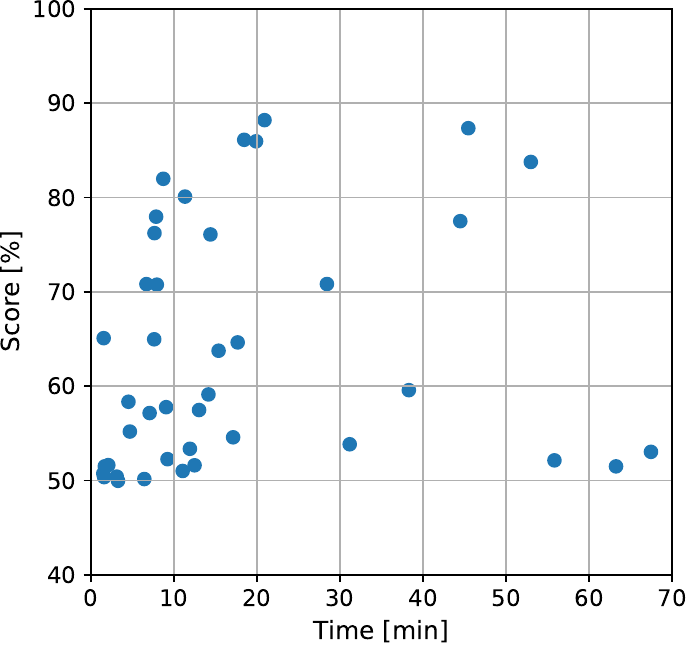}
    \includegraphics[width=0.45\linewidth, page=2]{fig3.pdf}
    \caption{Results of all the submitted algorithms: Score vs Time (left) and Accuracy on Test-set 1 vs Accuracy on Test-set 2 (right).}
    \label{fig:results_accuracy2}
\end{figure}

In Fig.\ref{fig:results_accuracy2} we present some additional analyses on all the submitted algorithms.
The left box aims at understanding how much does computational complexity (measured by the execution time to process 10,000 images) impact on the final score.
Interestingly, there is only a weak correlation between computation effort and performance, with methods that achieve the same very high score (around 90\%) with very different execution times.
The right box, instead, shows the results of each method on Test-set 1 and Test-set 2. 
In this case, a strong correlation is observed: if an algorithm performs well/bad on Test-set 1, this same happens on Test-set 2, even if the datasets do not overlap and are completely separated in terms of generating models.

\begin{figure*}
    \centering
    \begin{tabular}{ccc}
    {\small Test-set 1} &  {\small Test-set 2} & \\[-3pt]
    \savecellbox{\includegraphics[width=0.4\linewidth,page=1]{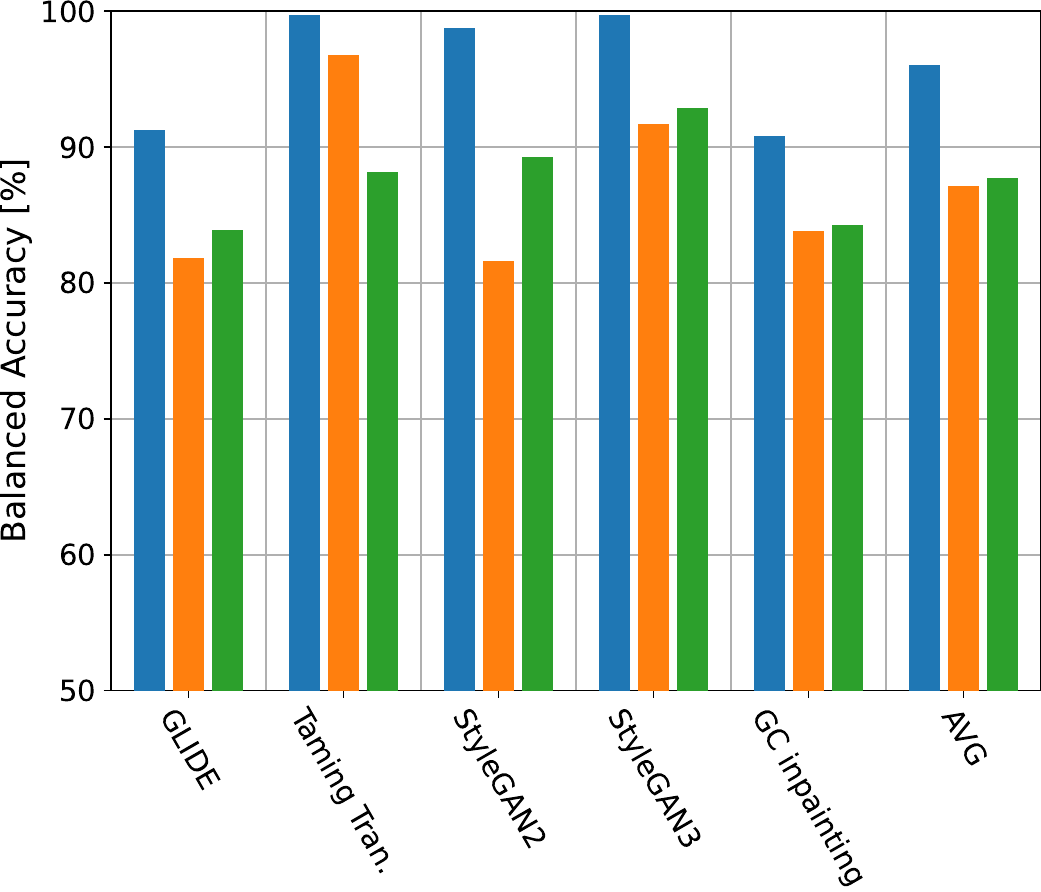}} &
    \savecellbox{\includegraphics[width=0.4\linewidth,page=2]{fig4.pdf}} &
    \savecellbox{\includegraphics[width=0.25\linewidth,page=3,angle=90]{fig4.pdf}}
     \\[-\rowheight] \printcelltop & \printcelltop & \printcelltop 
    \end{tabular}
    
    \caption{Balanced Accuracy of the three best performing methods on images from Test-set 1 and Test-set 2.}
    \label{fig:results_arc}
\end{figure*}

\begin{figure*}
    \centering
    \begin{tabular}{ccc}
    {\small Test-set 1} &  {\small Test-set 2} & \\[-3pt]
    \savecellbox{\includegraphics[width=0.4\linewidth,page=1]{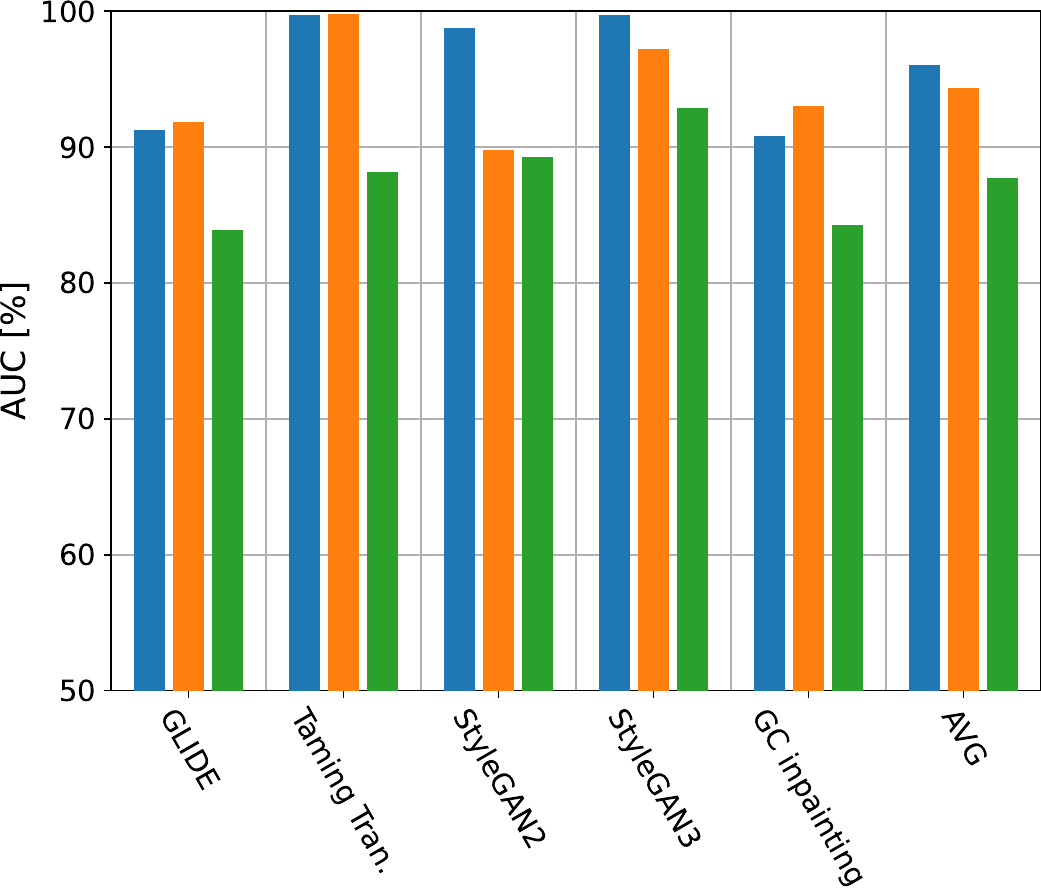}} &
    \savecellbox{\includegraphics[width=0.4\linewidth,page=2]{fig5.pdf}} &
    \savecellbox{\includegraphics[width=0.25\linewidth,page=3,angle=90]{fig5.pdf}}
     \\[-\rowheight] \printcelltop & \printcelltop & \printcelltop
    \end{tabular}
    
    \caption{Area Under the ROC (AUC) of the three best performing methods on images from Test-set 1 and Test-set 2.}
    \label{fig:results_auc}
\end{figure*}

Finally, in Fig.\ref{fig:results_arc}, we study in some more detail the performance of the three best performing techniques, reporting the balanced accuracy for each method on each dataset. 
For Test-set 1 (known models) the most difficult cases are those involving local manipulations. 
The same holds for Test-set 2 (unknown models) with the additional problem of images fully generated using diffusion models, where performance are on average lower than those obtained on images created by GANs. We also provide results in terms of AUC in Fig.\ref{fig:results_auc}. In this situation we can note that the first and second place reverse on Test-set 2, which underlines the importance to properly set the right threshold for the final decision. A proper choice of the validation set is indeed very important to carry out a good calibration.

\section{Highlights of the technical approaches}

In this Section we present an overview of the approaches proposed by all the participating teams for the challenge.
All proposed methods rely on learning-based approaches and train deep neural networks on a large dataset of real and synthetic images.
Many diverse architectures were considered: GoogLeNet, ResNet, Inception, Xception, DenseNet, EfficientNet, MobileNet, ResNeXt, ConvNeXt, and the more recent Vision Transformers.
The problem was often treated as a binary classification task (real vs fake) but some teams approached it as a multi-class classification problem
with the aim to increase the degrees of freedom for the predicting model and also to include an extra-class for unknown models.

In order to properly capture the forensic traces that distinguish pristine images from generated ones the networks considered multiple inputs, not just the RGB image. 
Indeed, it is well known that generators 
fail to accurately reproduce the natural correlation among color bands \cite{Li2020detection} and also that 
the upsampling operation routinely performed in most generative models gives rise to distinctive spectral peaks in the Fourier domain \cite{Zhang2019detecting}. 
Therefore, some solutions considered as input the image represented in different color spaces, i.e. HSV and YCbCr, or computed the co-occurrence matrices on the color channels. 
Moreover, to exploit frequency-based features, two-stream networks have been adopted, using features extracted from the Fourier analysis in the second stream.
A two-branch network was also used to work both on local and global features, that are fused by means of an attention module as done in \cite{Ju2022fusing}.
In general, attention mechanisms have been included in several solutions. 
Likewise, the ensembling of multiple networks was largely used to increase diversity and boost performance. 
Different aggregation strategies have been pursued with the aim to generalize to unseen models and favour decisions towards the real-image class, as proposed in \cite{Mandelli2022detecting}.

The majority of the teams trained their networks on the data made available for the challenge, 
however some of them increased this dataset by generating additional synthetic images using new generative models, such as other architectures based on GANs and new ones based on Diffusion models. Of course, including more generators during training helped to improve the performance, even if some approaches were able to obtain good generalization ability even adding few more models.
In addition, augmentation was always carried out to increase diversity and improve generalization. 
Beyond standard operations, like image flipping, cropping, resizing, and rotation, most teams used augmentation based on gaussian blurring and JPEG compression, found to be especially helpful in the literature \cite{Wang2020}, 
but also changes of saturation, contrast, brightness, as well as cutmix and random cutout.

\section{Finalists}
The final phase of the 2022 IEEE VIP Cup took place at ICIP in Bordeaux, on October 16, 2022. 
In the following, we describe the three finalist teams listed according to their final ranking: FAU Erlangen-Nürnberg (first place), Megatron (second place), Sherlock (third place). We will also present some details on their technical approach. 

\vspace{2mm}
\noindent
{\bf FAU Erlangen-Nürnberg}
\begin{itemize}
    \item {\em Affiliation}: Friedrich-Alexander-Universität Erlangen-Nürnberg, Germany
    \item {\em Supervisor}: Christian Riess
    \item {\em Tutor}:      Anatol Maier
    \item {\em Students}:   Vinzenz Dewor, Luca Beetz, ChangGeng Drewes, and Tobias Gessler
    \item {\em Technical approach}: an ensemble of vision transformers pre-trained on Imagenet-21k and fine-tuned on a large dataset of 400,000 images. 
                            To extract generalizable features, a procedure based on weighted random sampling was adopted during training aimed at balancing the data distribution.
                            Models included during training are the five known technique StyleGAN2 \cite{karras2020analyzing}, StyleGAN3 \cite{karras2021alias}, GLIDE \cite{nichol2021glide}, Taming Transformers \cite{esser2021taming} and inpainted images with Gated Convolution \cite{yu2019free}. In addition, images generated using DALL·E \cite{ramesh2021zero} and VQGAN \cite{esser2021taming} have been used.
\end{itemize}

\noindent
{\bf Megatron}
\begin{itemize}
    \item {\em Affiliation}: Bangladesh University of Engineering and Technology, Bangladesh
    \item {\em Supervisor}: Shaikh Anowarul Fattah
    \item {\em Students}:   Md Awsafur Rahman, Bishmoy Paul, Najibul Haque Sarker, and Zaber Ibn Abdul Hakim
    \item {\em Technical approach}: a multi-class classification scheme and an ensemble of convolutional neural networks and Transformer based architectures.
                            An extra-class is introduced to detect synthetic images coming from unknown models. 
                            Knowledge distillation and test-time augmentation are also included in the proposed solution.
                            The training set includes, beyond the five known techniques, additional images coming from the following  generators: ProGAN \cite{karras2018progressive}, ProjectedGAN \cite{sauer2021projected}, CycleGAN \cite{isola2017unpaired}, DDPM \cite{ho2020denoising}, Diffusion-GAN \cite{wang2022diffusiongan}, Stable Diffusion \cite{stablediffusion2022}, Denoising Diffusion GAN \cite{xiao2022tackling}, and GauGAN \cite{park2019semantic}.
\end{itemize}

\noindent
{\bf Sherlock}
\begin{itemize}
    \item {\em Affiliation}: Bangladesh University of Engineering and Technology, Bangladesh
    \item {\em Supervisor}: Mohammad Ariful Haque
    \item {\em Students}:   Fazle Rabbi, Asif Quadir, Indrojit Sarkar, Shahriar Kabir Nahin, Sawradip Saha, and Sanjay Acharjee
    \item {\em Technical approach}: a two-branch convolutional neural network that takes as input features extracted in the spatial and in the Fourier domain. 
                            The adopted architectures are EfficientNet-b7 and MobileNet-v3. 
                            In addition, strong augmentation is performed, which includes also cut-mix beyond standard operations.
                            During training only the five known generation techniques have been considered.
\end{itemize}

\section{Conclusions}

This article described the 2022 VIP Cup that took place last October at ICIP. The aim of the competition was to foster research on the detection of synthetic images, 
in particular, focusing on images generated using the recent diffusion models \cite{dhariwal2021diffusion,rombach2022high,balaji2022ediff,ramesh2022hierarchical}. 
These architectures have shown an impressive ability to generate images guided by textual descriptions or pilot sketches and there is very limited work on their detection \cite{corvi2022detection, Sha2022fake, ricker2022towards}. 
Below we want to highlight the main take-home messages that emerged from the technical solutions developed in this competition:
\begin{itemize}
    \item most performing models are pre-trained very deep networks that relied on a large dataset of real and synthetic images coming from several different generators. Indeed, increasing diversity during training was a key aspect of the best approaches;
    \item augmentation represents a fundamental step to make the model more robust to post-processing operations and make it work in realistic scenarios;
    \item generalization is still a main issue in synthetic image detection. In particular, it has been observed that one main problem is how to set the correct threshold in the more challenging scenario of unseen generators during training;
    \item the detection task can benefit of the attribution, which aims at identifying the model that was used for synthetic generation. 
\end{itemize}
We believe that the availability of the dataset\footnote{https://github.com/grip-unina/DMimageDetection} created during the challenge can stimulate the research on synthetic image detection and motivate other researchers to work in this interesting field. The advancements in generative AI make the distinction between real and fake very thin and it is very important to push the community to continuously search for effective solutions \cite{barni2023information}. In particular, the VIP Cup has shown the need to develop models that can be used in the wild to detect synthetic images generated by new architectures, such as the recent diffusion models. In this respect, it is important the design of explainable methods that can highlight which are the forensic artifacts the detector is exploiting \cite{corvi2023intriguing}.
We hope that more and more methods will be published in the research community and that will be inspired by the challenge proposed in the 2022 IEEE VIP Cup at ICIP.

\section{Acknowledgment}

The organizers would like to express their gratitude to all participating teams, to the local organizers at 2022 ICIP for hosting the VIP Cup, and to the IEEE Signal Processing Society Membership Board for the continuous support.
Special thanks go to Riccardo Corvi and Raffaele Mazza from University Federico II of Naples who helped to build the datasets. 
The authors want also to acknowledge the projects that support this research: 
DISCOVER within the SemaFor program funded by DARPA under agreement number FA8750-20-2-1004, 
Horizon Europe vera.ai funded by the European Union, Grant Agreement number 101070093, 
a TUM-IAS Hans Fischer Senior Fellowship,
PREMIER funded by the Italian Ministry of Education, University, and Research  within the PRIN 2017 program. This work is also funded by FCT/MCTES through national funds and when applicable co-funded EU funds under the projects UIDB/50008/2020 and LA/P/0109/2020.

\bibliographystyle{IEEEtran}
\bibliography{IEEEabrv,ref}

\begin{thebibliography}{10}
\providecommand{\url}[1]{#1}
\csname url@samestyle\endcsname
\providecommand{\newblock}{\relax}
\providecommand{\bibinfo}[2]{#2}
\providecommand{\BIBentrySTDinterwordspacing}{\spaceskip=0pt\relax}
\providecommand{\BIBentryALTinterwordstretchfactor}{4}
\providecommand{\BIBentryALTinterwordspacing}{\spaceskip=\fontdimen2\font plus
\BIBentryALTinterwordstretchfactor\fontdimen3\font minus
  \fontdimen4\font\relax}
\providecommand{\BIBforeignlanguage}[2]{{%
\expandafter\ifx\csname l@#1\endcsname\relax
\typeout{** WARNING: IEEEtran.bst: No hyphenation pattern has been}%
\typeout{** loaded for the language `#1'. Using the pattern for}%
\typeout{** the default language instead.}%
\else
\language=\csname l@#1\endcsname
\fi
#2}}
\providecommand{\BIBdecl}{\relax}
\BIBdecl

\bibitem{theguardian}
A.~Mahdawi, ``{Nonconsensual deepfake porn is an emergency that is ruining
  lives},''
  \url{https://www.theguardian.com/commentisfree/2023/apr/01/ai-deepfake-porn-fake-images},
  2023.

\bibitem{theverge}
J.~Vincent, ``{After deepfakes go viral, AI image generator Midjourney stops
  free trials citing ‘abuse’},''
  \url{https://www.theverge.com/2023/3/30/23662940/deepfake-viral-ai-misinformation-midjourney-stops-free-trials},
  2023.

\bibitem{Verdoliva2020media}
L.~Verdoliva, ``Media forensics and deepfakes: an overview,'' \emph{IEEE
  Journal of Selected Topics in Signal Processing}, vol.~14, no.~5, pp.
  910--932, 2020.

\bibitem{karras2019style}
T.~Karras, S.~Laine, and T.~Aila, ``A style-based generator architecture for
  generative adversarial networks,'' in \emph{IEEE/CVF Conference on Computer
  Vision and Pattern Recognition}, 2019, pp. 4396--4405.

\bibitem{nichol2021glide}
A.~Q. Nichol, P.~Dhariwal, A.~Ramesh, P.~Shyam, P.~Mishkin, B.~Mcgrew,
  I.~Sutskever, and M.~Chen, ``{GLIDE}: Towards photorealistic image generation
  and editing with text-guided diffusion models,'' in \emph{International
  Conference on Machine Learning}.\hskip 1em plus 0.5em minus 0.4em\relax PMLR,
  2022, pp. 16\,784--16\,804.

\bibitem{Dayma_DALLE_Mini_2021}
\BIBentryALTinterwordspacing
B.~Dayma, S.~Patil, P.~Cuenca, K.~Saifullah, T.~Abraham, P.~Lê~Khàc,
  L.~Melas, and R.~Ghosh, ``{DALL-E Mini},'' 2021. [Online]. Available:
  \url{https://github.com/borisdayma/dalle-mini}
\BIBentrySTDinterwordspacing

\bibitem{rombach2022high}
R.~Rombach, A.~Blattmann, D.~Lorenz, P.~Esser, and B.~Ommer, ``High-resolution
  image synthesis with latent diffusion models,'' in \emph{IEEE/CVF Conference
  on Computer Vision and Pattern Recognition}, 2022, pp. 10\,684--10\,695.

\bibitem{balaji2022ediff}
Y.~Balaji, S.~Nah, X.~Huang, A.~Vahdat, J.~Song, K.~Kreis, M.~Aittala, T.~Aila,
  S.~Laine, B.~Catanzaro, T.~Karras, and M.-Y. Liu, ``{eDiff-I: Text-to-Image
  Diffusion Models with Ensemble of Expert Denoisers},'' \emph{arXiv preprint
  arXiv:2211.01324}, 2022.

\bibitem{Gragnaniello2021GAN}
D.~Gragnaniello, D.~Cozzolino, F.~Marra, G.~Poggi, and L.~Verdoliva, ``Are gan
  generated images easy to detect? a critical analysis of the
  state-of-the-art,'' in \emph{IEEE International Conference on Multimedia and
  Expo}, 2021, pp. 1--6.

\bibitem{esser2021taming}
P.~Esser, R.~Rombach, and B.~Ommer, ``Taming transformers for high-resolution
  image synthesis,'' in \emph{IEEE/CVF Conference on Computer Vision and
  Pattern Recognition}, 2021, pp. 12\,873--12\,883.

\bibitem{karras2020analyzing}
T.~Karras, S.~Laine, M.~Aittala, J.~Hellsten, J.~Lehtinen, and T.~Aila,
  ``{Analyzing and improving the image quality of StyleGAN},'' in
  \emph{IEEE/CVF Conference on Computer Vision and Pattern Recognition}, 2020,
  pp. 8110--8119.

\bibitem{karras2021alias}
T.~Karras, M.~Aittala, S.~Laine, E.~H{\"a}rk{\"o}nen, J.~Hellsten, J.~Lehtinen,
  and T.~Aila, ``Alias-free generative adversarial networks,'' \emph{Advances
  in Neural Information Processing Systems}, vol.~34, pp. 852--863, 2021.

\bibitem{yu2019free}
J.~Yu, Z.~Lin, J.~Yang, X.~Shen, X.~Lu, and T.~S. Huang, ``Free-form image
  inpainting with gated convolution,'' in \emph{IEEE/CVF International
  Conference on Computer Vision}, 2019, pp. 4471--4480.

\bibitem{brock2018large}
A.~Brock, J.~Donahue, and K.~Simonyan, ``{Large Scale GAN Training for High
  Fidelity Natural Image Synthesis},'' in \emph{International Conference on
  Learning Representations}, 2019.

\bibitem{dhariwal2021diffusion}
P.~Dhariwal and A.~Nichol, ``{Diffusion models beat GANs on image synthesis},''
  \emph{Advances in Neural Information Processing Systems}, vol.~34, pp.
  8780--8794, 2021.

\bibitem{suvorov2022resolution}
R.~Suvorov, E.~Logacheva, A.~Mashikhin, A.~Remizova, A.~Ashukha, A.~Silvestrov,
  N.~Kong, H.~Goka, K.~Park, and V.~Lempitsky, ``Resolution-robust large mask
  inpainting with fourier convolutions,'' in \emph{IEEE/CVF Winter Conference
  on Applications of Computer Vision}, 2022, pp. 2149--2159.

\bibitem{deng2009imagenet}
J.~Deng, W.~Dong, R.~Socher, L.-J. Li, K.~Li, and L.~Fei-Fei, ``Imagenet: A
  large-scale hierarchical image database,'' in \emph{IEEE Conference on
  Computer Vision and Pattern Recognition}, 2009, pp. 248--255.

\bibitem{lin2014microsoft}
T.-Y. Lin, M.~Maire, S.~Belongie, J.~Hays, P.~Perona, D.~Ramanan,
  P.~Doll{\'a}r, and C.~L. Zitnick, ``{Microsoft COCO: Common objects in
  context},'' in \emph{European Conference on Computer Vision}, 2014, pp.
  740--755.

\bibitem{yu2015lsun}
F.~Yu, A.~Seff, Y.~Zhang, S.~Song, T.~Funkhouser, and J.~Xiao, ``{LSUN:
  Construction of a large-scale image dataset using deep learning with humans
  in the loop},'' \emph{arXiv preprint arXiv:1506.03365}, 2015.

\bibitem{Li2020detection}
H.~Li, B.~Li, S.~Tan, and J.~Huang, ``{Detection of deep network generated
  images using disparities in color components},'' \emph{Signal Processing},
  vol. 174, 2020.

\bibitem{Zhang2019detecting}
X.~Zhang, S.~Karaman, and S.-F. Chang, ``{Detecting and Simulating Artifacts in
  GAN Fake Images},'' in \emph{IEEE international Workshop on Information
  Forensics and Security}, 2019, pp. 1--6.

\bibitem{Ju2022fusing}
Y.~Ju, S.~Jia, L.~Ke, H.~Xue, K.~Nagano, and S.~Lyu, ``{Fusing Global and Local
  Features for Generalized AI-Synthesized Image Detection},'' \emph{IEEE
  International Conference on Image Processing}, pp. 3465--3469, 2022.

\bibitem{Mandelli2022detecting}
S.~Mandelli, N.~Bonettini, P.~Bestagini, and S.~Tubaro, ``{Detecting
  GAN-generated Images by Orthogonal Training of Multiple CNNs},'' in
  \emph{IEEE International Conference on Image Processing}, 2022, pp.
  3091--3095.

\bibitem{Wang2020}
S.-Y. Wang, O.~Wang, R.~Zhang, A.~Owens, and A.~Efros, ``{CNN-generated images
  are surprisingly easy to spot... for now},'' in \emph{IEEE/CVF Conference on
  Computer Vision and Pattern Recognition}, 2020, pp. 8692--8701.

\bibitem{ramesh2021zero}
A.~Ramesh, M.~Pavlov, G.~Goh, S.~Gray, C.~Voss, A.~Radford, M.~Chen, and
  I.~Sutskever, ``Zero-shot text-to-image generation,'' in \emph{International
  Conference on Machine Learning}.\hskip 1em plus 0.5em minus 0.4em\relax PMLR,
  2021, pp. 8821--8831.

\bibitem{karras2018progressive}
T.~Karras, T.~Aila, S.~Laine, and J.~Lehtinen, ``{Progressive Growing of GANs
  for Improved Quality, Stability, and Variation},'' in \emph{International
  Conference on Learning Representations}, 2018.

\bibitem{sauer2021projected}
A.~Sauer, K.~Chitta, J.~M{\"u}ller, and A.~Geiger, ``{Projected GANs converge
  faster},'' \emph{Advances in Neural Information Processing Systems}, 2021.

\bibitem{isola2017unpaired}
J.-Y. Zhu, T.~Park, P.~Isola, and A.~A. Efros, ``Unpaired image-to-image
  translation using cycle-consistent adversarial networks,'' in \emph{IEEE
  International Conference on Computer Vision}, 2017, pp. 2223--2232.

\bibitem{ho2020denoising}
J.~Ho, A.~Jain, and P.~Abbeel, ``{Denoising Diffusion Probabilistic Models},''
  \emph{Advances in Neural Information Processing Systems}, pp. 6840--6851,
  2020.

\bibitem{wang2022diffusiongan}
Z.~Wang, H.~Zheng, P.~He, W.~Chen, and M.~Zhou, ``{Diffusion-GAN: Training GANs
  with Diffusion},'' \emph{International Conference on Learning
  Representations}, 2023.

\bibitem{stablediffusion2022}
R.~Rombach, A.~Blattmann, D.~Lorenz, P.~Esser, and B.~Ommer, ``Stable
  diffusion,'' \url{https://github.com/CompVis/stable-diffusion}, 2022.

\bibitem{xiao2022tackling}
Z.~Xiao, K.~Kreis, and A.~Vahdat, ``{Tackling the Generative Learning Trilemma
  with Denoising Diffusion GANs},'' in \emph{International Conference on
  Learning Representations}, 2022.

\bibitem{park2019semantic}
T.~Park, M.-Y. Liu, T.-C. Wang, and J.-Y. Zhu, ``Semantic image synthesis with
  spatially-adaptive normalization,'' in \emph{IEEE/CVF Conference on Computer
  Vision and Pattern Recognition}, 2019, pp. 2332--2341.

\bibitem{ramesh2022hierarchical}
A.~Ramesh, P.~Dhariwal, A.~Nichol, C.~Chu, and M.~Chen, ``Hierarchical
  text-conditional image generation with clip latents,'' \emph{arXiv preprint
  arXiv:2204.06125v1}, 2022.

\bibitem{corvi2022detection}
R.~Corvi, D.~Cozzolino, G.~Zingarini, G.~Poggi, K.~Nagano, and L.~Verdoliva,
  ``On the detection of synthetic images generated by diffusion models,''
  \emph{IEEE International Conference on Acoustics, Speech and Signal
  Processing}, pp. 1--5, 2023.

\bibitem{Sha2022fake}
Z.~Sha, Z.~Li, N.~Yu, and Y.~Zhang, ``{DE-FAKE: Detection and Attribution of
  Fake Images Generated by Text-to-Image Diffusion Models},'' \emph{arXiv
  preprint arXiv:2210.06998}, 2022.

\bibitem{ricker2022towards}
J.~Ricker, S.~Damm, T.~Holz, and A.~Fischer, ``{Towards the Detection of
  Diffusion Model Deepfakes},'' \emph{arXiv preprint arXiv:2210.14571}, 2022.

\bibitem{barni2023information}
M.~Barni, P.~Campisi, E.~J. Delp, G.~Do{\"e}rr, J.~Fridrich, N.~Memon,
  F.~P{\'e}rez-Gonz{\'a}lez, A.~Rocha, L.~Verdoliva, and M.~Wu, ``{Information
  Forensics and Security: A Quarter Century Long Journey},'' \emph{IEEE Signal
  Processing Magazine}, 2023.

\bibitem{corvi2023intriguing}
R.~Corvi, D.~Cozzolino, G.~Poggi, K.~Nagano, and L.~Verdoliva, ``Intriguing
  properties of synthetic images: from generative adversarial networks to
  diffusion models,'' in \emph{IEEE Computer Vision and Pattern Recognition
  Workshops}, 2023, pp. 973--982.

\end{thebibliography}

\end{document}